\documentclass{article}

\usepackage[final, nonatbib]{neurips_2019}

\usepackage[utf8]{inputenc} 
\usepackage[T1]{fontenc}    
\usepackage{hyperref}       
\usepackage{url}            
\usepackage{booktabs}       
\usepackage{amsfonts}       
\usepackage{nicefrac}       
\usepackage{microtype}      
\usepackage{graphicx}
\usepackage[
backend=biber,
style=numeric-comp,
sorting=ynt
]{biblatex}
\usepackage{hyperref}
\hypersetup{
    colorlinks=true,
    linkcolor=blue,
    filecolor=magenta,      
    urlcolor=cyan,
}
 
\title{A Biologically Plausible Benchmark for\\Contextual Bandit Algorithms in Precision Oncology\\Using {\em in vitro} Data}

\author{%
  Niklas T. Rindtorff\thanks{Additional affiliation of Niklas Rindtorff, Nisarg Patel and MingYu Lu: Department of Biomedical informatics, Harvard Medical School, 10 Shattuck Street Boston, MA 02115}\\
  Department of Signaling \& Functional Genomics\\
  German Cancer Research Center\\
  INF 280, Heidelberg, 69120 Germany\\
  \texttt{n.rindtorff@dkfz.de}\\
  \And
  MingYu Lu\\
  Laboratory for Computational Physiology\\
  Massachusetts Institute of Technology\\
  \And
  Nisarg A. Patel\\
  Department of Oral and Maxillofacial Surgery\\
  University of California, San Francisco\\
  \And
  Huahua Zheng\\
  Department of Biostatistics\\
  Harvard T.H. Chan School of Public Health
  \And
  Alexander D'Amour\\
  Google Research\\
  355 Main Street, Cambridge, MA 02138 USA\\
}

\begin{document}

\maketitle

\begin{abstract}
  Precision oncology, the genetic sequencing of tumors to identify druggable targets, has emerged as the standard of care in the treatment of many cancers.
  Nonetheless, due to the pace of therapy development and variability in patient information, designing effective protocols for individual treatment assignment in a sample-efficient way remains a major challenge.
  One promising approach to this problem is to frame precision oncology treatment as a contextual bandit problem and to apply sequential decision-making algorithms designed to minimize regret in this setting.
  However, a clear prerequisite for considering this methodology in high-stakes clinical decisions is careful benchmarking to understand realistic costs and benefits.
  Here, we propose a benchmark dataset to evaluate contextual bandit algorithms based on real \textit{in vitro} drug response of approximately 900 cancer cell lines.
  Specifically, we curated a dataset of complete treatment responses for a subset of 7 treatments from prior \textit{in vitro} studies.
  This allows us to compute the regret of proposed decision policies using biologically plausible counterfactuals.
  We ran a suite of Bayesian bandit algorithms on our benchmark, and found that the methods accumulate less regret over a sequence of treatment assignment tasks than a rule-based baseline derived from current clinical practice.
  This effect was more pronounced when genomic information was included as context. 
  We expect this work to be a starting point for evaluation of both the unique structural requirements and ethical implications for real-world testing of bandit based clinical decision support.
\end{abstract}

\section{Introduction}
Precision oncology, the genetic sequencing of tumors to identify druggable targets, has quickly progressed as the standard of care in the treatment of many cancers \cite{Schwartzberg2017}. Here, targeted treatments show therapeutic activity in subsets of patients defined by tumor-specific genetic alterations, such as imatinib for chronic myelogenous leukemia. \cite{Blumenthal2010} However, assigning patients to adequate treatments remains challenging. Current practice applies published and approved therapeutic protocols that consider the patient's clinical characteristics, including the presence of (most often one) particular genetic mutation, to choose a therapeutic. For example, based on the presence of a single genetic variant, such as a BRAF V600E mutation (a gene involved in cell growth), a treatment decision can be made \cite{Hou2011}. However, this  limits the  ability to make high-confidence clinical decisions in a real-world scenario with a large number of both observable genetic data and treatment options to choose from. The implications for the practice of precision oncology are (I) a high selectivity: only 4.9\% of oncology patients are eligible for genome-targeted therapies with robust clinical evidence \cite{Marquart2018}, and subsequently (II) high compassionate use: the majority of oncology patients are left with limited treatment options outside of existing therapeutic protocols, of which off-label use does not contribute systematically to the development of new clinical evidence.

Given the nature of precision oncology, treatment assignment can be modeled as a contextual bandit problem with a patient's information informing the choice of treatment. 
In contrast to supervised learning on the one end and reinforcement learning on the other end, contextual bandit problems, especially when based on Thompson sampling, are a well-suited method for this task as it allow agents to explore new treatment options while ensuring that every action has a non-zero chance of being optimal \cite{Riquelme2018}. 
Put differently, Thompson sampling based agents would never make choices for a patient that are certainly non-optimal in order to improve decision making at a later time point, something that can not be excluded for more most full reinforcement learning algorithms.
While contextual bandit applications in oncology have been previously proposed \cite{Durand2015, Durand2018ContextualBF}, there are no established benchmarks to evaluate different algorithms, objectives, and state representations, due to a lack of biologically interpretable and complete observations of drug response in cancer. This is especially relevant as major ethical questions of how to balance the competing directives of individual utility and population utility remain.

Here we propose a benchmark for contextual bandits in precision oncology based on real \textit{in vitro} drug response of approximately 900 cancer cell lines \cite{Iorio2016}. For each cell line, mutation, copy number variation, and gene expression data is available to represent the sample's state. After defining rewards based on treatment response, we used all available algorithms implemented in the Bayesian bandit showdown project \cite{Riquelme2018,Snoek2015} to subsequently choose the best treatments for randomly selected cell lines. In addition, we defined a rule-based agent based on a set of current evidence-based therapeutic protocols to evaluate bandit performance and to include prior knowledge into the available state information during selected experiments.

\section{Methods}

\subsection{Benchmark Construction}

\begin{figure}
  \centering
  \includegraphics[width=13cm]{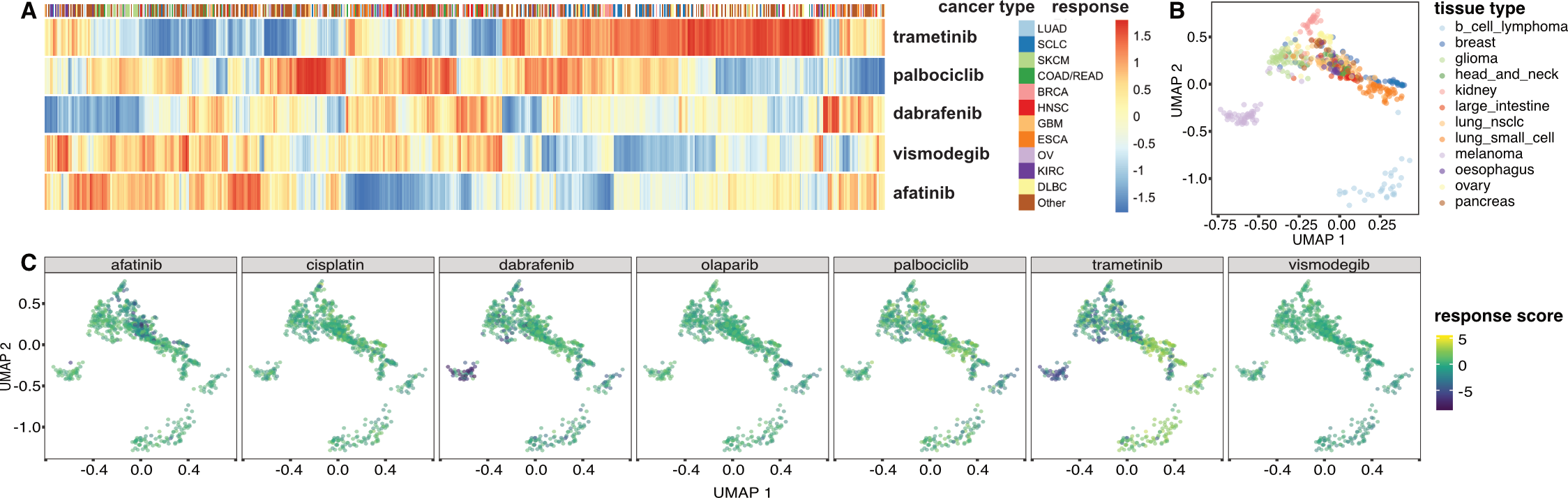}
  \caption{Benchmark Construction. \textbf{A} Drug vulnerability data for 896 cell lines and 5 drugs. Blue color corresponds to high drug sensitivity. Only 5 of 7 treatment choices are shown. 
  \textbf{B} 20-dimensional embedding of genomic information for all cancer cell lines included in the study. The first two dimensions of the UMAP embedding recovered tissue type differences between cancer cell lines.
  \textbf{C} The first two dimensions of the UMAP embedding did not completely recover differences in drug sensitivity}
  \label{fig:umap_tissue}
\end{figure}

We derived all molecular and drug sensitivity data from the Genomics of Drug Sensitivity in Cancer database, a public research repository described by Iorio et al \cite{Iorio2016} and available at \url{https://www.cancerrxgene.org}. In a first pre-processing step we focused on a subset of 7 drugs that are currently used in clinical practice. We log transformed the $IC50$ values and normalized them relative to the median $ln(IC50)$ across cell-lines for each drug. We used the resulting score to quantify drug response and calculate response-based rewards (Figure \ref{fig:umap_tissue}\textbf{A}).


In order to reduce the dimensionality of the state representation, we reduced 18523 cell-line specific features including scaled gene expression data, and binarized mutation and copy-number variant information into 20 dimensions by uniform manifold approximation and projection (UMAP) using default parameters. UMAP projected features recovered tissue types (Figure \ref{fig:umap_tissue}\textbf{B}) while not directly recovering overall drug sensitivities (Figure \ref{fig:umap_tissue}\textbf{C}).  

\begin{figure}
  \centering
  \includegraphics[width=13cm]{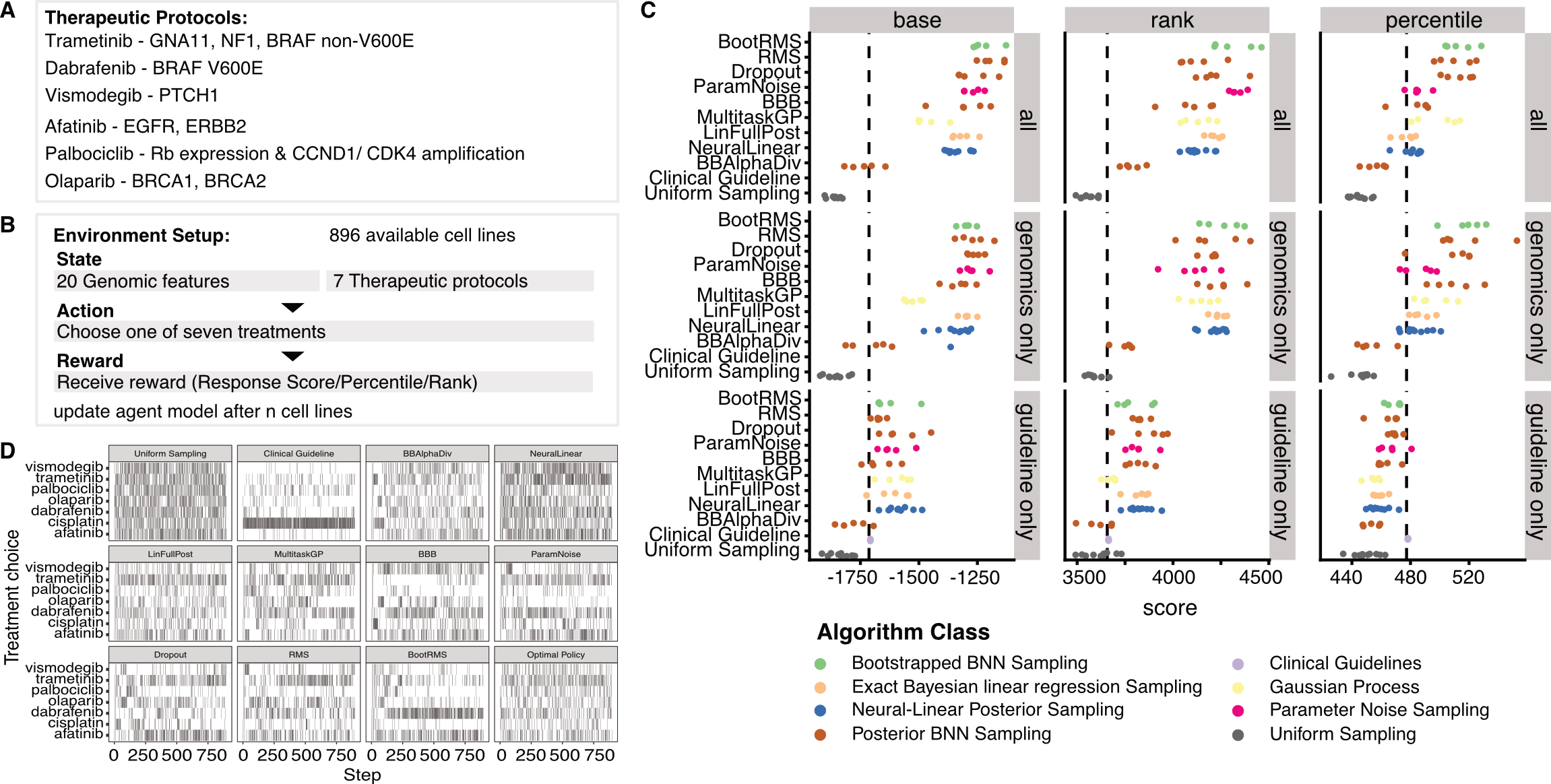}
  \caption{Contextual bandit experimental results. 
  \textbf{A} Rules derived from current therapeutic protocols which are used by the refernce "Clinical Guideline" agent.
  \textbf{B} Overview of the experimental setup. Both state representation and reward function are varied between experiments.
  \textbf{C} Overview of model performance across reward metrics and state definitions. Dashed line represents the performance of the rule-based agent.
  A higher score indicates lower regret and better performance.
  \textbf{D} Agent activity over time for different posterior distribution approximations. In this example, state was represented by both genetic information and clinical guidelines. In this instance, RMS was the most successful model.}
  \label{fig:response_score}
\end{figure}

Next we manually curated therapeutic protocols based on \href{https://pct.mdanderson.org/home}{current clinical evidence}, \href{http://oncokb.org/}{established} and \href{https://moalmanac.org/}{recent} databases, as well as \href{https://clinicaltrials.gov/ct2/show/NCT02465060}{trial protocols} with selected simplifications: (I) we excluded any protocols involving combination treatments, (II) we excluded any protocols that are based on the presence of oncogenic gene-fusions as they were not included in our dataset, (III) in analogy to a basket trials, we did not include tissue type restrictions into any protocols. Cell lines that did not qualify for any of the curated treatments were assigned to be treated with Cisplatin, an established chemotherapeutic used, among others, for treatment of cancers of unknown progeny (Figure \ref{fig:response_score}\textbf{A}).  

\subsection{Contextual Bandit Formulation}

We followed the definition of the contextual bandit problem as described in \cite{Riquelme2018}. The algorithm assigns treatments to units sequentially. At a time $t = 1, \cdots, n$ the algorithm takes as input a context corresponding to the next unit  $X_t$ (e.g., a cell line's projected genetic data). The algorithm selects one of k actions $a_t$ (e.g., one of 7 available treatments). A reward $r_t = r_t ( X_t , a_t )$ is then generated and returned. At the end, the cumulative reward for the algorithm is defined as $r = \sum_{t=1}^{n}r^t$. The goal is to maximize cumulative reward, and thus minimize the cumulative regret, defined as $R_A = E[r - r^*]$, where $r^*$ is the cumulative reward of the optimal policy (i.e., the policy that always selects the ideal treatment given the context).

Similar to the study in \cite{Riquelme2018}, we exclusively examined the performance of decision making via Thompson sampling. In each round of Thompson Sampling, parameters, $\theta_t$, are sampled from the posterior distribution $\pi_{t-1}$ given all previous observations. Using these parameters and the current context $X_t$, an action $a_t$ is chosen to optimize the expected regret $E[r_t| \theta_t, a_t, x_t]$ according to an internal model. 

In this study, we examined the effect of (I) different state representations (II) different reward functions and (III) different posterior distribution approximations \cite{Riquelme2018} on performance. We evaluated three different state representations by including only genetic features, only rule-based recommendations or both datatypes in the state. Thus, in total, the state was represented by up to 27 features (20 UMAP + 7 recommendations). Further we defined three different reward metrics: 

\begin{itemize}
    \item subtract the lowest drug response score (the strongest response) from the response score of the selected drug.
    \item rank the drugs by drug response score in ascending order. The best drug will be ranked 7, while the least active drug will be ranked 1.
    \item for each drug, we map its response score to its distribution over cell lines and use the percentile as reward. 
\end{itemize}

For posterior approximation, we consider all of the Bayesian bandit algorithms included in \cite{Riquelme2018} with default parameters.
These included uniform sampling, Bayesian linear regression, Gaussian Processes, stochastic variational inference, and several neural-network based approximations.
A full listing of methods and their hyperparameters are included in the appendix.

For each state representation, reward function and all posterior approximators, we ran 100 epochs with 512 as the batch size for deep Bayesian network training to obtain the final results. We repeated each experiment with 5 independent random seeds, thus generating 5 random patient sequences to go through. We did not define a separate validation dataset to measure agent performance, as commonly done in methods such as cross-validation, because the UMAP representation was learned on the complete dataset, leading to an overestimation of agent performance.

All experiments were run in python 3.6 using modified code from the \textbf{Deep Bayesian Bandits Library}, including an additional rule-based "Clinical Guideline" agent that followed current therapeutic protocols and a logging function to export an agent's actions over all experiment steps.  

\section{Results}

Overall, our results suggest that contextual bandit algorithms show promise in the precision oncology setting.
In our experiments, contextual bandits methods were able to leverage genomic information to consistently achieve substantially lower regret than both uniform random allocation and rule-based clinical guidelines \ref{fig:response_score}\textbf{B}.
This main result is shown in Figure \ref{fig:response_score}\textbf{C} with an exemplary plot visualising agent activity \ref{fig:response_score}\textbf{D}.
Specifically, in a baseline experiment where each algorithm was only given the information needed to implement clinical guidelines, all agents out-performed uniform random allocation, and had comparable performance to the rule-based reference agent (bottom row).
This was to be expected, as clinical guidelines have already been tuned to take advantage of this information.
However, when genomic information was available, most of the contextual bandit algorithms were able to improve on the rule-based protocol significantly (top and middle rows).
Providing both genomic information and guideline input in the state information did not further improve model performance in most cases (top row vs middle row). 

These results were generally robust across reward definitions (columns), although the percentile-based reward showed the weakest results.
Of note, three Neural Network based algorithms, bootstrapped-, greedy and Dropout, consistently scored higher rewards compared to linear methods or Gaussian Processes.

\section{Discussion}

In summary, we state that genomics based assignment mechanisms in precision oncology programs can be framed as a contextual bandit problem. When provided with a representation of genomic information, contextual bandit agents can outperform simplified abstractions of current clinical standards based on \textit{in vitro} drug response data. Among the most successful agents were bootstrapped or dropout-based dense neural networks.

This study has several limitations including: (I) \textit{In vitro} drug response data of cancer models has limited transferability into a clinical context although recovering a considerable portion of clinically established genetic predictors of drug response \cite{Iorio2016}, (II) The response scores are on average lower in treatments vs. reference agents, (III) We reduced the dimensionality of available genomic data without dedicated learning of a shared multi-omics embedding, for example as described in \cite{Simidjievski719542} (IV) Cisplatin is a limited reference treatment for all considered cancer types. 

We decided to reduce the dimensionality of the available feature space in order to reduce model complexity and increase training efficiency. We chose UMAP for this purpose as it recovers both global- and local structure of the dataset. As mentioned before, we believe that this dataset does not only offer a benchmark for machine learning based treatment assignment for cancer, but also action-oriented mulitomic feature representation of this disease. 

In the future, we plan to address the limitations above and validate our findings in alternative \textit{in vitro} and \textit{in vivo} drug response datasets \cite{Gao2015}, which were measured by perturbing cancer cell lines, patient-derived organoids or xenografts. Clinical outcome data, although valuable, does not lend itself directly for benchmarking, as not all available treatments have been observed for every patient and thus no ideal policy beyond the standard of care is known. Nevertheless, we plan to validate out finding by analyzing agent behaviour for action-patterns that correspond to current clinical best practices.
In addition, we plan to measure the impact of certain genomic information types on model performance, for example by using only the available information provided by current genetic testing services.

We would like to stimulate an open discussion about the limitations and potential benefits of bandit-guided treatment assignments in precision oncology programs to minimize collective treatment regret.

\section{Code and Data availability}
All code and data can be accessed in this \href{https://github.com/NiklasTR/oncoassign}{repository} 
or the following \href{https://drive.google.com/open?id=1zjHghTA3Rm3ljEBDcN8HULxYBlOrvIYp}{directory}. 

\newpage

\printbibliography

\appendix
\section{Full Listing of Bayesian Bandit Algorithms}

Here we list the full suite of Bayesian bandit algorithms that we evaluated with our benchmark.
\begin{itemize}
    \item Uniform Sampling (Takes each action at random with equal probability)
    \item Bayesian linear (Noise prior $a_0 = 6$, $b_0 = 6$. Ridge prior $\lambda$ = 0.25)
    \item Neural Linear (Noise prior $a_0 = 3$, $b_0 = 3$. Ridge prior $\lambda = 0.25$. Based on RMS2 net)
    \item Neural Greedy (Greedy NN approach with fixed learning rate ($\gamma = 0.01$))
    \item Dropout (Dropout with probability p = 0.8. Based on RMS3 net)
    \item Parameter-Noise (Initial noise $\sigma$ = 0.01, and level $\epsilon$ = 0.01. Based on RMS2 net)
    \item Bootstrapped Networks (Bootstrapped with $q = 5$ models, and $p = 0.85$. Based on RMS3 net)
    \item Stochastic Variational Inference (BayesByBackprop with noise $\sigma = 0.1$)
    \item Expectation-Propagation (Alpha Divergences BB $\alpha$-divergence with $\alpha$ = 0.1, noise $\sigma$ = 0.1, $K = 10,$ prior var $\sigma^2_0 = 0.1$.)
    \item RMS2 net (Learning rate decays, and it is reset every training period)
    \item RMS3 net (Learning rate decays, and it is not reset at all. Starts at $\gamma = 1 $)
\end{itemize}

\end{document}